\pdfoutput=1

\documentclass[11pt]{article}

\usepackage{EMNLP2022}

\usepackage{times}
\usepackage{latexsym}

\usepackage[T1]{fontenc}

\usepackage[utf8]{inputenc}

\usepackage{microtype}
\definecolor{na}{gray}{0.9}
\usepackage{color}
\usepackage{makecell}
\usepackage{multirow}
\usepackage{xspace}
\usepackage{colortbl}
\usepackage{amsfonts}
\usepackage{inconsolata}
\usepackage{epsfig,graphics,float}
\usepackage{caption}
\usepackage{booktabs}
\usepackage{amsmath}
\usepackage{mathtools}
\usepackage{sidecap}
\usepackage{multirow}
\usepackage{epsfig,graphics,float}
\usepackage{todonotes}
\usepackage[ruled,vlined]{algorithm2e}
\usepackage{subfig,balance,lineno}
\usepackage{float}
\usepackage{verbatim}

\usepackage{wrapfig}

\newcommand{\sysname}{{\tt AdaMix}}

\title{AdaMix: Mixture-of-Adaptations for Parameter-efficient Model Tuning}

\author{Yaqing Wang$^{\S}$, Sahaj Agarwal$^\diamond{}$, Subhabrata Mukherjee$^\dagger{}$, Xiaodong Liu$^\dagger{}$, \\\textbf{Jing Gao}$^{\S}$, \textbf{Ahmed Hassan Awadallah}$^\dagger{}$, \textbf{Jianfeng Gao}$^\dagger{}$ \\
  $^{\S}$Purdue University,
   $^\diamond{}$Microsoft,
  $^\dagger{}$Microsoft Research\\
  \texttt{\{wang5075, jinggao\}@purdue.edu},\\ \texttt{\{sahagar, submukhe, xiaodl, hassanam, jfgao\}@microsoft.com}}

\begin{document}
\maketitle
\begin{abstract}
Standard fine-tuning of large pre-trained language models (PLMs) for downstream tasks requires updating hundreds of millions to billions of parameters, and storing a large copy of the PLM weights for every task resulting in increased cost for storing, sharing and serving the models. To address this, parameter-efficient fine-tuning (PEFT) techniques were introduced where small trainable components are injected in the PLM and updated during fine-tuning. 
We propose {\sysname} as a general PEFT method that tunes a mixture of adaptation modules -- given the underlying PEFT method of choice -- introduced in each Transformer layer while keeping most of the PLM weights frozen. For instance, AdaMix can leverage a mixture of adapters like Houlsby~\cite{houlsby} or a mixture of low rank decomposition matrices like LoRA~\cite{hu2021lora} to improve downstream task performance over the corresponding PEFT methods for fully supervised and few-shot NLU and NLG tasks. Further, we design AdaMix such that it matches the same computational cost and the number of tunable parameters as the underlying PEFT method. By only tuning $0.1-0.2\%$ of PLM parameters, we show that AdaMix  outperforms SOTA parameter-efficient fine-tuning and full model fine-tuning for both NLU and NLG tasks. Code and models are made available at \href{https://aka.ms/AdaMix}{https://aka.ms/AdaMix}.




\end{abstract}

\section{Introduction}

\begin{figure}[htbp]
 	\centering
  	\includegraphics[width=2.2in]{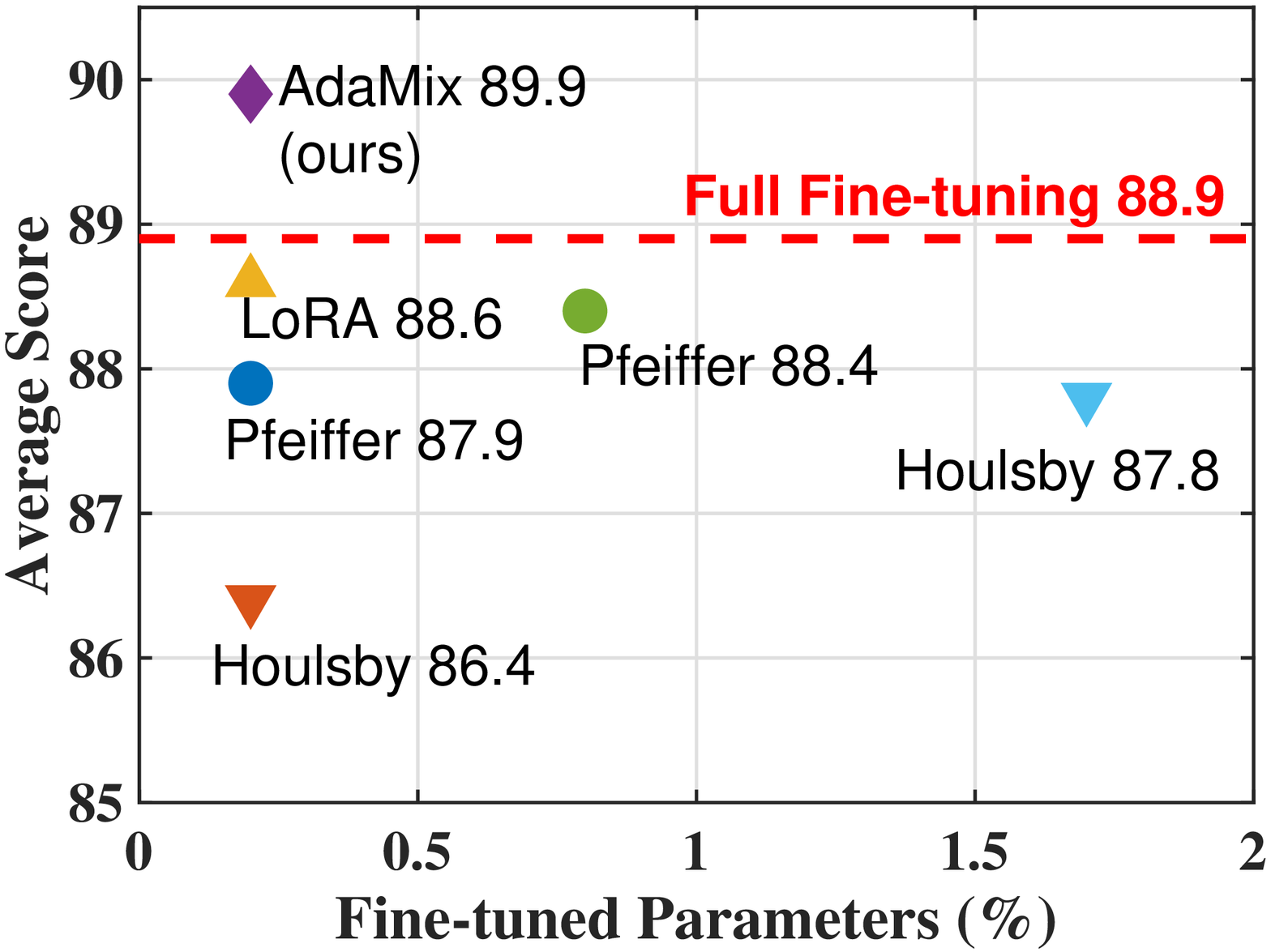}
 	  	\caption{Performance of different parameter-efficient fine-tuning methods on GLUE development set with RoBERTa-large encoder following a setup similar to~\cite{houlsby} for fair comparison. We report the performance of Pfeiffer~\cite{pfeiffer2021adapter}, Houlsby~\cite{houlsby} and LoRA~\cite{hu2021lora} with their default number of fine-tuned parameters as well as the number of fine-tuned parameters used in {\sysname} with a mixture of adaptations . Red dash shows the performance of full model fine-tuning.}
 	 	 	\label{fig:openfig}
 	 	 	\vspace{-1.6em}
\end{figure}

 Standard fine-tuning of large pre-trained language models (PLMs)~\citep{BERT, RoBERTa,brown2020language,  raffel2019exploring} to downstream tasks requires updating all model parameters. Given the ever-increasing size of PLMs (e.g., $175$ billion parameters for GPT-3~\cite{brown2020language}  and $530$ billion parameters for MT-NLG~\cite{smith2022using}), even the fine-tuning step becomes expensive as it requires storing a full copy of model weights for every task. 
To address these challenges, recent works have developed parameter-efficient fine-tuning (PEFT) techniques. These approaches typically underperform standard full model fine-tuning, but significantly reduce the number of trainable parameters. There are many varieties of PEFT methods, including prefix-tuning~\citep{prefix} and prompt-tuning~\citep{prompt_tuning} to condition frozen language models via natural language task descriptions, low dimensional projections using adapters~\citep{houlsby,pfeiffer2020AdapterHub,pfeiffer2021adapter} and more recently using low-rank approximation~\citep{hu2021lora}. Figure~\ref{fig:openfig} shows the performance of some popular PEFT methods with varying number of tunable parameters. We observe a significant performance gap with respect to full model tuning where all PLM parameters are updated. 

In this paper, we present {\sysname}, a mixture of adaptation modules approach, and show that it outperforms SOTA PEFT methods and also full model fine-tuning while tuning only $0.1-0.2\%$ of PLM parameters.

In contrast to traditional PEFT methods that use a single adaptation module in every Transformer layer, {\sysname} uses several adaptation modules that learn multiple views of the given task. In order to design this mixture of adaptations, we take inspiration from sparsely-activated mixture-of-experts (MoE) models. In traditional dense models (e.g., BERT~\citep{BERT}, GPT-3~\citep{brown2020language}), all model weights are activated for every input example.  %
MoE models induce sparsity by activating only a subset of the model weights for each incoming input. 

Consider adapters~\citep{houlsby}, one of the most popular PEFT techniques, to illustrate our method. A feedforward layer (FFN) is introduced to {\em down-project} the hidden representation to a low dimension $d$ (also called the bottleneck dimension) followed by another {\em up-project} FFN to match the dimensionality of the next layer. Instead of using a single adapter, we introduce multiple project-up and project-down FFNs in each Transformer layer. We route input examples to one of the project-up and one of the project-down FFN's resulting in the same amount of computational cost (FLOPs) as that of using a single adapter. For methods like LoRA~\cite{hu2021lora}, that decomposes the gradient of pre-trained weights into low-rank matrices ($A$ and $B$), we introduce multiple low-rank decompositions and route the input examples to them similar to adapters.


We discuss different routing mechanism and show that stochastic routing yields good performance while eliminating the need for introducing any additional parameters for module selection. To alleviate training instability that may arise from the randomness in selecting different adaptation modules in different training steps, we leverage consistency regularization and the sharing of adaptation modules during stochastic routing. 

The introduction of multiple adaptation modules results in an increased number of adaptation parameters. This does not increase computational cost but increases storage cost. To address this, we develop a merging mechanism to combine weights from different adaptation modules to a single module in each Transformer layer. This allows us to keep the number of adaptation parameters the same as that of a single adaptation module. Our merging mechanism is inspired by model weight averaging model soups~\citep{modelsoup} and multi BERTs~\citep{sellam2022the}. Weight averaging of models with different random initialization has been shown to improve model performance in recent works~\citep{matena2021merging, neyshabur2020being, frankle2020linear} that show the optimized models to lie in the same basin of error landscape. While the above works are geared towards fine-tuning independent models, we extend this idea to parameter-efficient fine-tuning with randomly initialized adaptation modules and a frozen language model. 

Overall, our work makes the following contributions:

\noindent {\bf (a)} We develop a new method {\sysname} as a mixture of adaptations for parameter-efficient fine-tuning (PEFT) of large language models. Given any PEFT method of choice like adapters and low-rank decompositions, {\sysname} improves downstream task performance over the underlying PEFT method. 

\noindent {\bf (b)} {\sysname} is trained with stochastic routing and adaptation module merging to retain the same computational cost (e.g., FLOPs, \#tunable adaptation parameters) and benefits of the underlying PEFT method. To better understand how {\sysname} works, we demonstrate its strong connections to Bayesian Neural Networks and model ensembling.

\noindent {\bf (c)} By tuning only $0.1-0.2\%$ of a pre-trained language model’s parameters, {\sysname} is the first PEFT method to outperform full model fine-tuning methods 
for all NLU tasks on GLUE, and outperforms other competing methods for NLG and few-shot NLU tasks.

\noindent\textbf{Practical benefits of PEFT methods.} The most significant benefit of PEFT methods comes from the reduction in memory and storage usage. For a Transformer, the VRAM consumption can be significantly reduced as we do not need to keep track of optimizer states for the frozen parameters. PEFT methods also allow multiple tasks to share the same copy of the full (frozen) PLM. Hence, the storage cost for introducing a new task can be reduced by up to 444x (from 355MB to 0.8MB with RoBERTa-large encoder in our setting).



{\em We present background on Mixture-of-Experts (MoE) and adapters in Section~\ref{sec:background} of Appendix.}

\begin{figure*}[htb!]
 		\centering
	\includegraphics[width=0.8\textwidth]{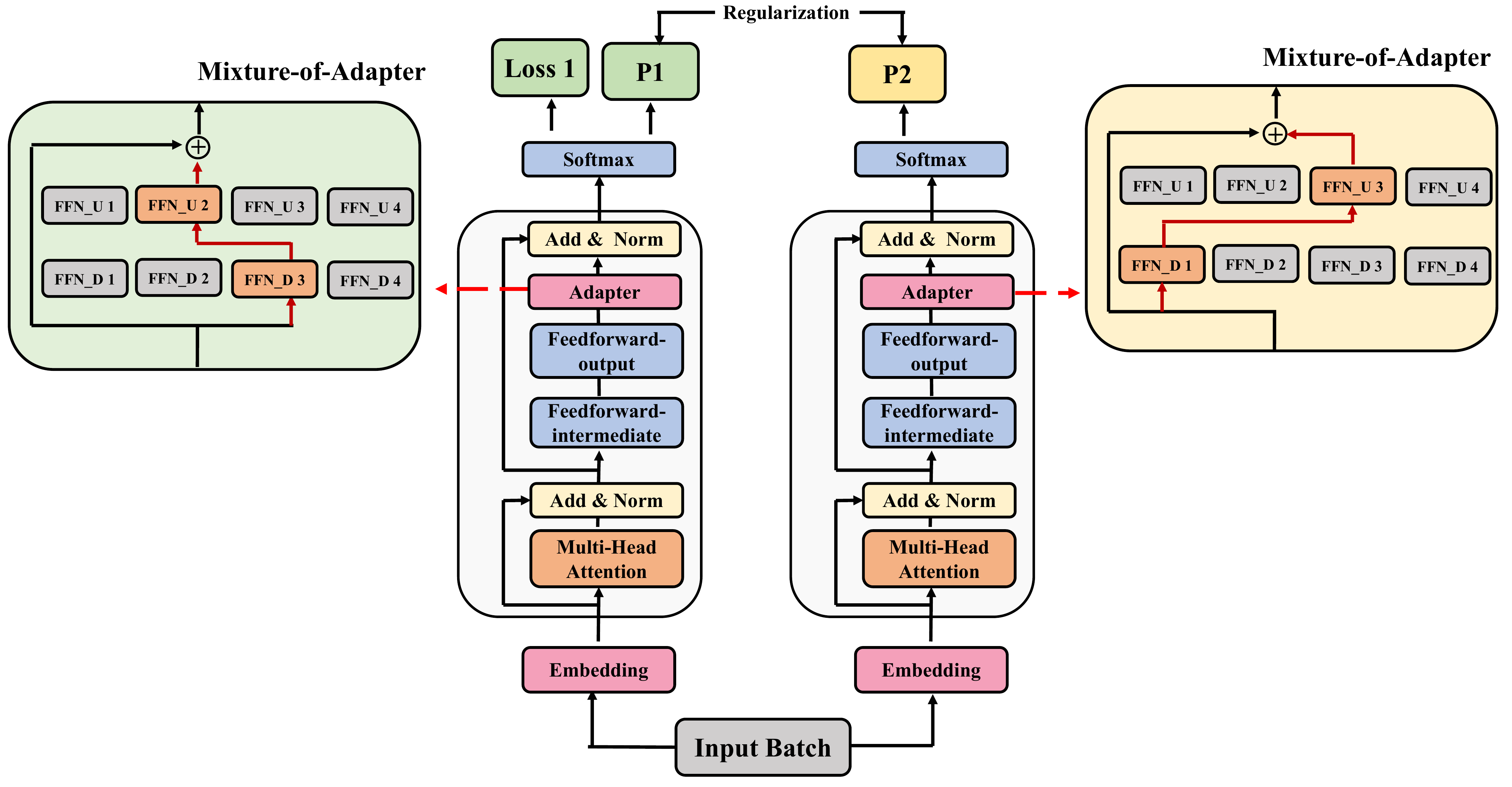}
 	  	\caption{Mixture-of-Adaptations ({\sysname}) with adapters~\cite{houlsby} as the underlying PEFT mechanism. For illustration, we show $M=4$ adaptation modules consisting of feedforward up ($\text{FFN\_U}$) feedforward down ($\text{FFN\_D})$ projection matrices. The above block shown for one Transformer layer is repeated across all the layers. {\sysname} stochastically routes instances from an input batch via randomly selected adaptation modules resulting in FLOPs match to a single module with consistency regularization and parameter sharing. Adaptation merging (Figure~\ref{fig:inference}) collapses multiple modules to match single-module parameters in each layer.}
 	 	 	\label{fig:framework}
 
\end{figure*}

\section{Background}
\label{sec:background}
\subsection{Mixture-of-Experts}


The objective of sparsely-activated model design is to support conditional computation and increase the parameter count of neural models like Transformers while keeping the floating point operations (FLOPs) for each input example constant. Mixture-of-Experts (MoE) Transformer models~\citep{shazeer2017outrageously, fedus2021switch,lepikhin2020gshard,zuo2021taming} achieve this by using $N$ feed-forward networks (FFN), namely ``experts" denoted as $\mathbb{E}_{i=1}^N$, each with its own set of learnable weights that compute different representations of an input token $x$ based on context. In order to sparsify the network to keep the FLOPs constant, there is an additional gating network $\mathbb{G}$ whose output is a sparse $N$-dimensional vector to route each token via a few of these experts. Note that, a sparse model with $N=1$ corresponding to only one FFN layer in each Transformer block collapses to the traditional dense model. 

Consider $x_s$ as the input token representation in the $s^{th}$ position to the MOE layer comprising of the $\{\mathbb{E}\}_{i=1}^N$ expert FFNs. Also, consider $w^{in}_i$ and $w^{out}_i$ to be the input and output projection matrices for $i^{th}$ expert. Expert output  $\mathbb{E}_i(x_s)$ is given by:

\begin{equation}
    \mathbb{E}_i(x_s) = w^{out}_i \cdot GeLU (w^{in}_i \cdot x_s)
\end{equation}

\noindent Consider $\mathbb{G}(x_s)$ to be output of the gating network. Output of the sparse MoE layer is given by:

\begin{equation}
\label{eq:moe}
    h(x_s)=\sum_i \mathbb{G}(x_s)_i\ \mathbb{E}_i(x_s) 
\end{equation}

\noindent where $\mathbb{G}(x_s)_i$ the $i^{th}$ logit of the output of $\mathbb{G}(x_s)$ denotes the probability of selecting expert $\mathbb{E}_i$. 


In order to keep the number of FLOPs in the sparse Transformer to be the same as that of a dense one, the gating mechanism can be constrained to route each token to only one expert FFN, i.e. $\sum_i \mathbb{G}_t(x_s)_i = 1$.

\subsection{Adapters}
\label{subsec:adapter}

\begin{figure}
 	\centering
  	\includegraphics[width=1.8in]{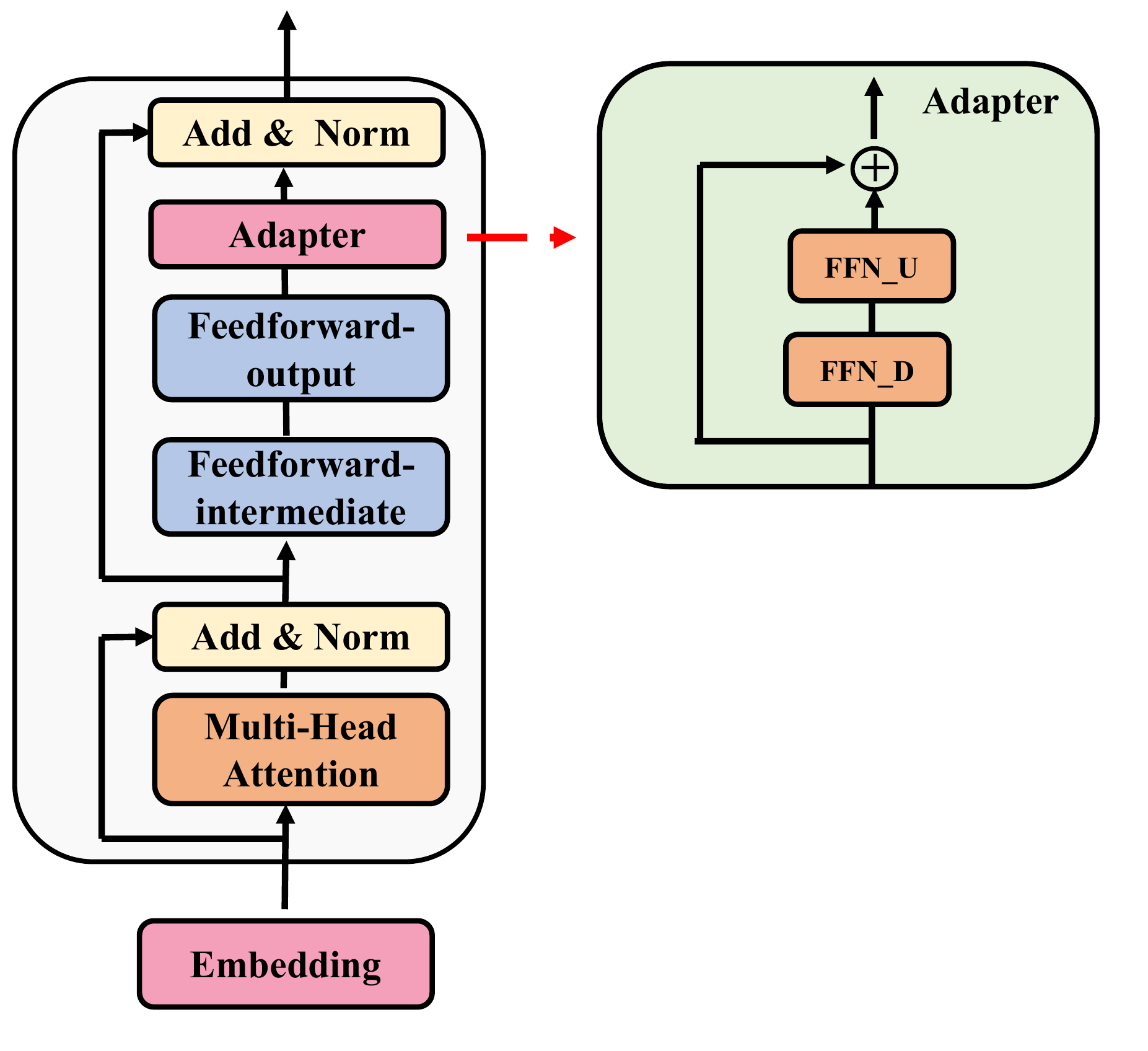}
 	  	\caption{Conventional adapter design in standard Transformer architecture.}
 	 	 	\label{fig:houlsby-adapter}
\end{figure}

The predominant methodology for task adaptation is to tune all of the trainable parameters of the PLMs for every task. This raises significant resource challenges both during training and deployment. A recent study~\citep{aghajanyan2020intrinsic} shows that PLMs have a low instrinsic dimension that can match the performance of the full parameter space. 

To adapt PLMs for downstream tasks with a small number of parameters, adapters~\citep{houlsby} have recently been introduced as an alternative approach for lightweight tuning. 

The adapter tuning strategy judiciously introduces new parameters into the original PLMs. During fine-tuning, only the adapter parameters are updated while keeping the remaining parameters of the PLM frozen. Adapters usually consist of two fully connected layers as shown in Figure~\ref{fig:houlsby-adapter}, where the adapter layer uses a down projection $\mathcal{W}^{down} \in \mathcal{R}^{d \times r}$ to project input representation $x$ to a low dimensional space $r$ (referred as the bottleneck dimension) with $d$ being the model dimension, followed by
a nonlinear activation function $f(\cdot)$, and a up-projection with $\mathcal{W}^{up} \in \mathcal{R}^{r \times d}$ to project the low-dimensional features back to the original dimension.  
The adapters are further surrounded by residual connections.

Given the above adapter design with parameters $\psi$, the dataset $\mathcal{D}_K$, a pre-trained language model encoder $enc$ with parameters $\Theta_{\mathrm{PLM}}$, where $\Theta_{\mathrm{PLM}} \gg \psi$, we want to perform the following optimization for efficient model adaptation:
\begin{equation}
\label{eq:opt}
\psi \leftarrow argmin_{\psi}\  \mathcal{L} (\mathcal{D}_k; \Theta_{\mathrm{PLM}}, \psi)
\end{equation}

\section{Mixture-of-Adaptations}

Consider a set of $M$ adaptation modules injected in each Transformer layer, where $A_{ij}: i \in \{1 \cdots L\}, j \in \{1 \cdots M\}$ represents the $j^{th}$ adaptation module in the $i^{th}$ Transformer layer. For illustration, we will consider adapters~\cite{houlsby} as the underlying parameter-efficient fine-tuning (PEFT) mechanism as a running example. Similar principles can be used for other PEFT mechanism like LoRA~\cite{hu2021lora} for low-rank decompositions as we show in experiments.

We adopt the popularly used Transformer architecture~\cite{vaswani2017attention} consisting of $L$ repeated Transformer blocks, 
%
%
%
where each block consists of a self-attention sub-layer, a fully connected feed-forward network (FFN) and residual connections around the sub-layers followed by layer normalization. Each adaptation module $A_{ij}$ corresponding to the adapters~\cite{houlsby} consists of a feedforward up $\mathcal{W}^{up}_{ij}$ and a feedforward down $\mathcal{W}^{down}_{ij}$ projection matrices.


\subsection{Routing Policy} 
Recent work like THOR~\cite{zuo2021taming} has demonstrated stochastic routing policy like random routing to work as well as classical routing mechanism like Switch routing~\cite{fedus2021switch} with the following benefits. Since input examples are randomly routed to different experts, there is no requirement for additional load balancing as each expert has an equal opportunity of being activated simplifying the framework. Further, there are no added parameters, and therefore no additional computation, at the Switch layer for expert selection. The latter is particularly important in our setting for parameter-efficient fine-tuning to keep the parameters and FLOPs the same as that of a single adaptation module. To analyze the working of {\sysname}, we demonstrate connections to stochastic routing and model weight averaging to Bayesian Neural Networks and model ensembling in Section~\ref{subsec:BNN}.

In the stochastic routing policy for {\sysname} with adapters, at any training step, we randomly select a pair of feedforward up and feedforward down projection matrices in the $i^{th}$ Transformer layer as $A_i=\{\mathcal{W}^{up}_{ij}, \mathcal{W}^{down}_{ik}\}$ and $B_i=\{\mathcal{W}^{up}_{ij'},  \mathcal{W}^{down}_{ik'}\}$ respectively. Given this selection of adaptation modules $A_i$ and $B_i$ in each Transformer layer in every step, all the inputs in a given batch are processed through the same set of modules. Given an input representation $x$ in a given Transformer layer, the above pair of modules perform the following transformations:

\begin{equation}
\setlength{\abovedisplayskip}{0pt}
\setlength{\belowdisplayskip}{0pt}
        x \leftarrow x + f(x \cdot \mathcal{W}^{down})\cdot \mathcal{W}^{up}
\end{equation}

Such stochastic routing enables adaptation modules to learn different transformations during training and obtain multiple views of the task. However, this also creates a challenge on which modules to use during inference due to random routing protocol during training. We address this challenge with the following two techniques that further allow us to collapse adaptation modules and obtain the same computational cost (FLOPs, \#tunable adaptation parameters) as that of a single module.

\subsection{Consistency regularization}
Consider $\mathcal{A}=\{A_{i=1}^{L}\}$ and $\mathcal{B}=\{ B_{i=1}^{L}\}$ to be the sets of adaptation modules (e.g., projection matrices) activated during two stochastic forward passes through the network for an input $x$ across $L$ layers of the Transformer. The objective of consistency regularization is to enable the adaptation modules to share information and prevent divergence. To this end, we add the following consistency loss as a regularizer to the task-specific optimization loss:
%
%
\setlength{\abovedisplayskip}{0pt}
\setlength{\belowdisplayskip}{0pt}
\begin{multline}
\small
\label{eq:loss}
    \mathcal{L} = - \bigg(\sum_{c=1}^C \mathcal{I}(x,c)\ \log\text{softmax}(z_c^{\mathcal{A}}(x))  +\\\frac{1}{2}\big(\mathcal{KL}(z_{(.)}^{\mathcal{A}}(x) || z_{(.)}^{\mathcal{B}}(x)) + \mathcal{KL}(z_{(.)}^{\mathcal{B}}(x) || z_{(.)}^{\mathcal{A}}(x))\bigg)
\end{multline}

\noindent where $\mathcal{I}(x, c)$ is a binary indicator ($0$ or $1$) if class label $c$ is the correct classification for $x$ and $z_{(.)}^{\mathcal{A}}(x)$ and $z_{(.)}^{\mathcal{B}}(x)$ are the predicted logits while routing through two sets of adaptation modules $\mathcal{A}$ and $\mathcal{B}$ respectively with $\mathcal{KL}$ denoting the Kullback-Leibler divergence. $x$ is the input representation from the PLM with frozen parameters and only the parameters of modules $\{\mathcal{W}^{up}, \mathcal{W}^{down}\}$ are updated during training.

\begin{figure}
 	\centering
  	\includegraphics[width=1\linewidth]{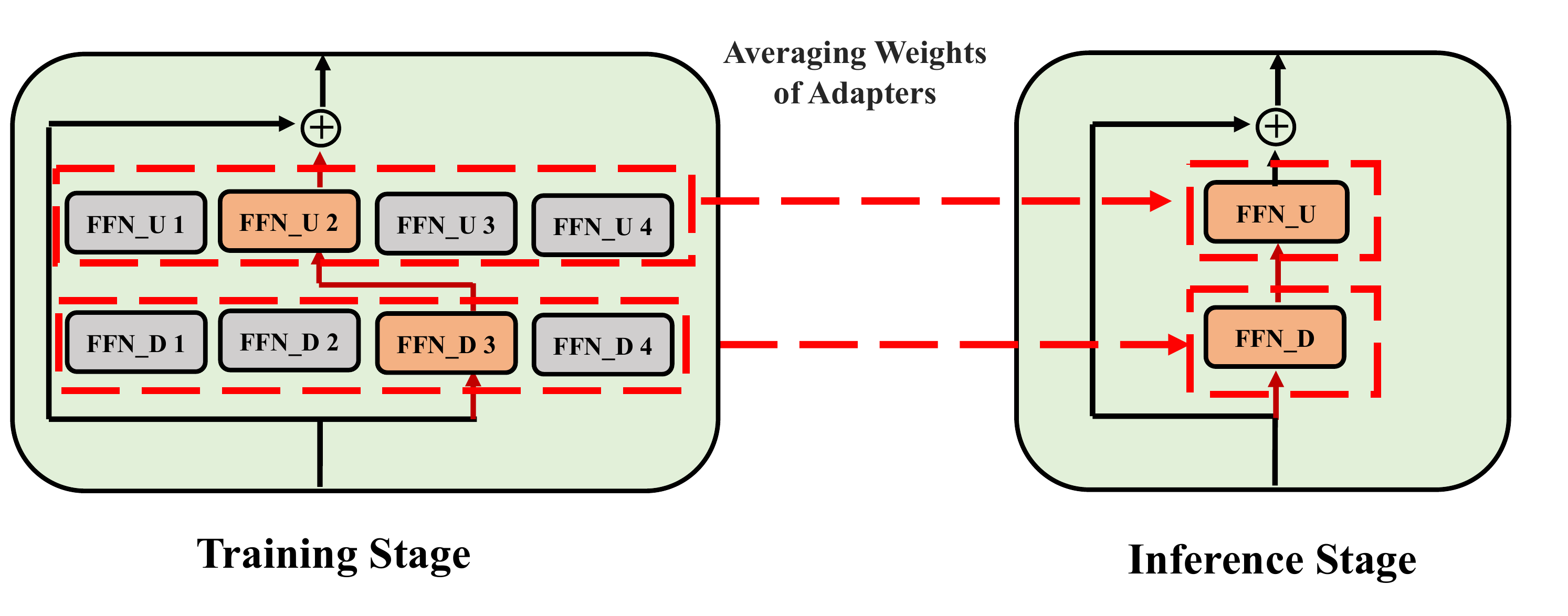}
 	  	\caption{Stochastic routing during training activates different adaptation modules to have multiple views of the task with FLOPs match to a single module. Merging weights of the adaptation modules ($\{\text{FFN}\_\text{U}_i\}, \{\text{FFN}\_\text{D}_i\}: i \in \{1 \cdots 4\}$) by averaging preserves improved performance with parameter match to a single-module.}
 	 	 	\label{fig:inference}
 	 	
\end{figure}

\subsection{Adaptation module merging}
While the above regularization mitigates inconsistency in random module selection during inference, it still results in increased serving cost to host several adaptation modules. Prior works in fine-tuning language models for downstream tasks have shown improved performance on averaging the weights of different models fine-tuned with different random seeds outperforming a single fine-tuned model. Recent work~\cite{modelsoup} has also shown that differently fine-tuned models from the same initialization lie in the same error basin motivating the use of weight aggregation for robust task summarization. We adopt and extend prior techniques for language model fine-tuning to our parameter-efficient training of multi-view adaptation modules.

In contrast to the aforementioned techniques like stochastic routing and consistency regularization that are applied at the training phase, we employ adaptation merging {\bf only during inference}. Given a set of adaptation modules, $\mathcal{W}^{up}_{ij}$ and $\mathcal{W}^{down}_{ik}$ for $i \in \{1 \cdots L\}$ and $\{j,k\} \in \{1 \cdots M\}$, we simply average the weights of all the corresponding modules (e.g., project-up or project-down matrices) in every Transformer layer to collapse to a single module $\{\mathcal{W'}^{up}_{i}, \mathcal{W'}^{down}_{i}\}$, where:

\begin{equation}
\setlength{\abovedisplayskip}{0pt}
\setlength{\belowdisplayskip}{0pt}
\small
\label{eq:merging}
    \mathcal{W'}^{up}_{i} \leftarrow \frac{1}{M} \sum_{j=1}^M \mathcal{W}^{up}_{ij} \qquad
    \mathcal{W'}^{down}_{i} \leftarrow  \frac{1}{M} \sum_{j=1}^M \mathcal{W}^{down}_{ij}
\end{equation}

\subsection{Adaptation module sharing} 
While stochastic routing to multi-view adaptation modules increases the model capacity, it can also impact downstream tasks with less amounts of labeled data for tuning several sets of adaptation {modules}. To address this challenge, we use another mechanism to share some of the adaption modules (e.g., project-down {\em or} the project-up operations) {to improve training efficiency}. In the standard setting for adapters, we share only the feedforward projection-up matrices i.e., $\mathcal{W}^{up}_{ij} = \mathcal{W}^{up}_i$. We investigate these design choices via ablation studies in our experiments in Section~\ref{sec:ablation} {and Section~\ref{appendix_sec:ablation} in Appendix}.



\subsection{Connection to Bayesian Neural Networks and Model Ensembling}
\label{subsec:BNN}

Bayesian Neural Network (BNN)~\citep{DBLP:journals/corr/GalG15} replaces a deterministic model's weight parameters by a distribution over the parameters. For inference, BNN averages over all the possible weights, also referred to as marginalization. 
Consider $f^{\mathcal{W}(x)} \in \mathbb{R}^d$ to be the $d-$dimensional output of such a neural network where the model likelihood is given by $p(y|f^{\mathcal{W}(x)})$. In our setting, $\mathcal{W}=\langle \mathcal{W}^{up}, \mathcal{W}^{down} \rangle$ along with frozen PLM parameters that are dropped from the notation for simplicity. For classification, we can further apply a softmax likelihood to the output to obtain: $P(y=c|x,W)=softmax(f^{\mathcal{W}(x)})$. 
Given an instance $x$, the probability distribution over the classes is given by marginalization over the posterior distribution as: $p(y=c|x) = \int_{\mathcal{W}} p(y=c|f^{\mathcal{W}(x)})p(\mathcal{W}|X,Y)d\mathcal{W}$.


This requires averaging over all possible model weights, which is intractable in practice. Therefore, several approximation methods have been developed based on variational inference methods and stochastic regularization techniques using dropouts. In this work, we leverage another stochastic regularization in the form of random routing.
%
Here, the objective is to find a surrogate distribution $q_\theta(w)$ in a tractable family of distributions that can replace the true model posterior that is hard to compute. The ideal surrogate is identified by minimizing the Kullback-Leibler (KL) divergence between the candidate and the true posterior. 


Consider $q_\theta(\mathcal{W})$ to be the stochastic routing policy which samples $T$ masked model weights $\{\widetilde{\mathcal{W}_t}\}_{t=1}^T \sim q_\theta(\mathcal{W})$. For classification tasks, the approximate posterior can be now obtained by Monte-Carlo integration~\cite{pmlr-v70-gal17a} as:

\begin{align}
\setlength{\abovedisplayskip}{0pt}
\setlength{\belowdisplayskip}{0pt}
\begin{split}
\label{eq:posterior}
    p(y=c|x) &\approx p(y=c|f^\mathcal{W}(x))q_\theta(\mathcal{W})d\mathcal{W} \\
    &\approx \frac{1}{T}\ \sum_{t=1}^{T} p(y=c|f^{\widetilde{\mathcal{W}_t}}(x))\\ &= \frac{1}{T}\ \sum_{t=1}^{T} softmax(f^{\widetilde{\mathcal{W}_t}}(x))
\end{split}
\end{align}

However, computing the approximate posterior above in our setting requires storing all the stochastic model weights $\mathcal{W}_t(x)$ which increases the serving cost during inference. To reduce this cost, we resort to the other technique for weight averaging via adaptation module merging during inference.


Let $\mathcal{L}^{AM}_{\mathcal{W}} = \mathbb{E}_{x,y} \mathcal{L}(softmax(f^{\widetilde{\mathcal{W}}}(x), y)$ denote the expected loss with merging of the stochastic adaptation weights with $\widetilde{\mathcal{W}} = \frac{1}{T} \sum_t \widetilde{\mathcal{W}_t}$ (from Equation~\ref{eq:merging}) and $\mathcal{L}$ denoting the cross-entropy loss.  Consider $\mathcal{L}^{Ens}_{\mathcal{W}} = \mathbb{E}_{x,y} \mathcal{L}(\frac{1}{T}\ \sum_{t=1}^{T} softmax(f^{\widetilde{\mathcal{W}_t}}(x), y))$ denote the expected loss from logit-level stochastic model ensembling (from Equation~\ref{eq:posterior}). 

Prior work~\citep{modelsoup} shows that averaging the weights of multiple models fine-tuned with different hyper-parameters improves model performance. They analytically show the similarity in loss between weight-averaging ($\mathcal{L}^{AM}_{\mathcal{W}}$ in our setting) and logit-ensembling ($\mathcal{L}^{Ens}_{\mathcal{W}}$ in our setting) as a function of the flatness of the loss and confidence of the predictions. While the above analysis is geared towards averaging of multiple independently fine-tuned model weights, we can apply a similar analysis in our setting towards averaging of multiple stochastically obtained adaptation weights in obtaining a favorable loss $\mathcal{L}^{AM}_{\mathcal{W}}$. Further, adaptation merging reduces the serving cost during inference since we need to retain only one copy of the merged weights as opposed to logit-ensembling which requires copies of all the adaptation weights.

\begin{table*}[!t]
\small

	\begin{center}
	\resizebox{1.0\linewidth}{!}{
		\begin{tabular}{lr c cccccc  c c}
			\toprule \bf Model &   \#Param. &MNLI        &QNLI               &SST2          &QQP          &MRPC               &CoLA           &RTE        &STS-B&\bf Avg. \\ 
			                           &&Acc             &Acc             &Acc          &Acc            &Acc             &Mcc            &Acc            &Pearson    \\ \midrule
			Full Fine-tuning$^\dagger{}$ &355.0M&90.2&94.7&96.4& 92.2&90.9&68.0&86.6&\textbf{92.4}&88.9 \\ \midrule
			Pfeiffer Adapter$^\dagger{}$&3.0M &90.2&94.8&96.1&91.9&90.2&68.3&83.8&92.1 &88.4 \\ 
			Pfeiffer Adapter$^\dagger{}$&0.8M &90.5&94.8&96.6&91.7&89.7&67.8&80.1&91.9&87.9  \\ 
			Houlsby Adapter$^\dagger{}$&6.0M&89.9&94.7&96.2&92.1&88.7&66.5&83.4&91.0& 87.8  \\ 
			Houlsby Adapter$^\dagger{}$&0.8M&90.3&94.7&96.3&91.5&87.7&66.3&72.9&91.5&86.4  \\ 
			LoRA$^\dagger{}$&0.8M&90.6&94.8&96.2&91.6&90.2&68.2&85.2&92.3 &88.6 \\ 
			{\sysname} Adapter &0.8M &\textbf{90.9}&\textbf{95.4}&\textbf{97.1}&\textbf{92.3}&\textbf{91.9}&\textbf{70.2}&\textbf{89.2}&\textbf{92.4} &\textbf{89.9}\\
			\bottomrule

		\end{tabular}}
	\end{center}

	\caption{Results for NLU tasks on GLUE development set with \textbf{RoBERTa-large} encoder. The best result on each task is in \textbf{bold} and ``-'' denotes missing measure. {\sysname} with a mixture of adapters outperforms all competing methods as well as fully fine-tuned large model with only $0.23\%$ tunable parameters.$^\dagger{}$~denotes results reported from \cite{hu2021lora}. Mcc refers to Matthews correlation coefficient, and Pearson refers to Pearson correlation. 
	\#Param. denotes the number of tunable adaptation parameters used during inference.}
	\label{tab:roberta_glue_dev}

\end{table*}

\section{Experiments}

\subsection{Experimental Setup}
\label{subsec:setup}
\noindent\textbf{Dataset.} We perform experiments on a wide range of tasks including eight natural language understanding (NLU) tasks in the General Language Understanding Evaluation (GLUE) benchmark \cite{wang2019glue} and three natural language generation (NLG) tasks, namely, E2E~\cite{novikova2017e2e}, WebNLG~\cite{gardent2017webnlg} and DART~\cite{nan2020dart}. For the NLU and NLG tasks, we follow the same setup as~\cite{houlsby} and~\cite{prefix,hu2021lora}, respectively. 

\noindent\textbf{Baselines.}
We compare
{\sysname} to full model fine-tuning and several state-of-the-art parameter-efficient fine-tuning (PEFT) methods, namely, Pfeiffer Adapter~\cite{pfeiffer2021adapter}, Houlsby Adapter~\cite{houlsby}, BitFit~\cite{zaken2021bitfit}, Prefix-tuning~\cite{prefix}, UNIPELT~\cite{mao2021unipelt} and LoRA~\cite{hu2021lora}. We use BERT-base~\cite{BERT} and RoBERTa-large~\cite{RoBERTa} as encoders for NLU tasks (results in Table~\ref{tab:roberta_glue_dev} and Table~\ref{tab:bert_glue_dev}), and GPT-2~\cite{brown2020language} for NLG tasks (results in Table~\ref{tab:nlg}). 

\noindent{\bf AdaMix implementation details.} We implement {\sysname} in Pytorch and use Tesla V100 gpus for experiments with detailed hyper-parameter configurations presented in Section~\ref{sec:hyper} in Appendix. {\sysname} with adapters uses a dimension of $16$ and $48$ using BERT-base and RoBERTa-large encoders following the setup of~\cite{hu2021lora, mao2021unipelt} for fair comparison. {\sysname} with LoRA uses rank $r=4$ following the setup of~\cite{hu2021lora} to keep the same number of adaptation parameters during inference. The number of adaptation modules in {\sysname} is set to $4$ for all the tasks and encoders unless otherwise specified. The impact of adapter dimension and number of adaptation modules for NLU tasks are investigated in Table~\ref{tab:ablation_adapter_num} and \ref{tab:ablation_adapter_dim}. {\em For most of the experiments and ablation analysis, we report results from {\sysname} with adapters for NLU tasks. For demonstrating the generalizability of our framework, we report results from {\sysname} with LoRA~\cite{hu2021lora} as the underlying PEFT mechanism for NLG tasks.}

\subsection{Key Results}

\subsubsection{NLU Tasks}

Tables~\ref{tab:roberta_glue_dev} and \ref{tab:bert_glue_dev} show the performance comparison among PEFT  models with  RoBERTa-large and BERT-base encoders respectively. Fully fine-tuned RoBERTa-large and BERT-base provide the ceiling performance. We observe {\sysname} with a mixture-of-adapters to significantly outperform other state-of-the-art baselines on most tasks with different encoders. 
 {\sysname} with adapters is the only PEFT method which outperforms full model fine-tuning on all the tasks and on average score.

\begin{table}[!htb]
\small
	\begin{center}
		\begin{tabular}{lrc}
			\toprule \bf Model    &\#Param.     &\bf Avg. \\ 
			\midrule
			Full Fine-tuning$^\dagger{}$&   110M &82.7\\ \midrule
			Houlsby Adapter$^\dagger{}$&0.9M&83.0\\ 
			 BitFit$^\diamond$ & 0.1M& 82.3\\
			Prefix-tuning$^\dagger{}$&0.2M&82.1\\ 
			LoRA$^\dagger{}$&0.3M &82.2\\ 
			UNIPELT (AP)$^\dagger{}$&1.1M&83.1\\ 
			UNIPELT (APL)$^\dagger{}$&1.4M&83.5\\ 

				{{\sysname} Adapter}  &0.9M& \textbf{84.5}\\
				\bottomrule

		\end{tabular}
	\end{center}
	\caption{Results for NLU tasks on GLUE development set with \textbf{BERT-base} encoder and {\sysname} with a mixture-of-adapters. The best result on each task is in \textbf{bold}. $^\dagger{}$ and $^\diamond$ denote results reported from \cite{mao2021unipelt,zaken2021bitfit}. Detailed task-specific results are reported in Table~\ref{tab:app-bert_glue_dev} of Appendix. \#Param. refers to the number of tunable adaptation parameters during inference.
	}
	\label{tab:bert_glue_dev}
\end{table}

\subsubsection{NLG Tasks}

\begin{table*}[!t]
\small

	\begin{center}
		\begin{tabular}{lr c cccc}
			\toprule \bf Model &   \#Param. & BLEU & NIST & MET & ROUGE-L & CIDEr \\ \midrule
Full Fine-tuning$^\dagger{}$ & 354.92M & 68.2 & 8.62 & 46.2 & 71.0 & 2.47\\\midrule
Lin AdapterL$^\dagger{}$& 0.37M &66.3 &8.41& 45.0& 69.8& 2.40\\
Lin Adapter$^\dagger{}$& 11.09M& 68.9& 8.71& 46.1& 71.3& 2.47\\
Houlsby Adapter$^\dagger{}$& 11.09M & 67.3& 8.50& 46.0& 70.7& 2.44\\
FT$^{Top2^\dagger{}}$ & 25.19M & 68.1 & 8.59 & 46.0 & 70.8 & 2.41\\
PreLayer$^\dagger{}$ & 0.35M & 69.7 & 8.81 & 46.1 & 71.4 & 2.49\\
LoRA$^\dagger{}$ & 0.35M & 70.4 & 8.85 & {\bf 46.8} & 71.8 & 2.53\\\midrule
LoRA (repr.) & 0.35M & 69.8 & 8.77 & 46.6 & 71.8 & 2.52\\
{\sysname} Adapter & 0.42M & 69.8 & 8.75 & {\bf 46.8} & 71.9 & 2.52\\
{\sysname} LoRA & 0.35M & {\bf 71.0} & {\bf 8.89} & {\bf 46.8} & {\bf 72.2} & {\bf 2.54}\\
 \bottomrule

		\end{tabular}
	\end{center}
	\caption{Results on E2E NLG Challenge with GPT-2 medium backbone. Best result on each task is in \textbf{bold}. We report {\sysname} results with both adapters and LoRA as underlying PEFT method. {\sysname} outperforms all competing methods as well as fully fine-tuned large model with only $0.1\%$ tunable parameters.$^\dagger{}$~denotes results reported from \cite{hu2021lora} and repr. denotes reproduced results. \#Param. denotes the number of tunable adaptation parameters used during inference. Results on DART and WebNLG presented in Tables~\ref{tab:dart} and~\ref{tab:webnlg} in Appendix.}
	\label{tab:nlg}
\end{table*}

{\sysname} leverages mixture of adaptations to improve over underlying PEFT method as demonstrated in Table~\ref{tab:nlg} for E2E NLG i.e. {\sysname} with LoRA and {\sysname} with adapters outperform LoRA~\cite{hu2021lora} and adapters~\cite{houlsby} respectively. We report results on DART and WebNLG in Tables~\ref{tab:dart} and~\ref{tab:webnlg} in Appendix.

\begin{table}[!hbt]
\small

	\begin{center}
		\begin{tabular}{lr c }
			\toprule \bf Model &   \#Param. & BLEU\\\midrule 
Full Fine-tuning$^\dagger{}$ & 354.92M & 46.2 \\\midrule
Lin AdapterL$^\dagger{}$& 0.37M & 42.4\\ 
Lin Adapter$^\dagger{}$& 11.09M& 45.2\\ 
FT$^{Top2^\dagger{}}$ & 25.19M & 41.0\\ 
PrefLayer$^\dagger{}$ & 0.35M & 46.4\\ 
LoRA$^\dagger{}$ & 0.35M & 47.1\\ 
LoRA (repr.) & 0.35M & 47.35\\ 
{\sysname} Adapter & 0.42M & 47.72\\ 
{\sysname} LoRA & 0.35M & {\bf 47.86}\\ 
 \bottomrule
 
		\end{tabular}
	\end{center}
	\vspace*{-1em}
	\caption{Results on DART with GPT-2 backbone encoder. Best result on each task is in \textbf{bold}. We report {\sysname} results with both adapters and LoRA as underlying PEFT method. {\sysname} outperforms all competing methods as well as fully fine-tuned large model with only $0.1\%$ tunable parameters.$^\dagger{}$~denotes results reported from \cite{hu2021lora} and repr. denotes reproduced results. \#Param. denotes the number of tunable adaptation parameters used during inference.}
	\label{tab:dart}
\vspace*{-1em}
\end{table}

\begin{table}[!hbt]
\small

	\begin{center}
		\begin{tabular}{lr c }
			\toprule \bf Model &   \#Param. & BLEU\\ \midrule
Full Fine-tuning$^\dagger{}$ & 354.92M & 46.5\\\midrule
Lin AdapterL$^\dagger{}$& 0.37M & 50.2 \\
Lin Adapter$^\dagger{}$& 11.09M& 54.9 \\
FT$^{Top2^\dagger{}}$ & 25.19M & 36.0 \\
Prefix$^\dagger{}$ & 0.35M & 55.1 \\
LoRA$^\dagger{}$ & 0.35M & 55.3 \\
LoRA (repr.) & 0.35M & 55.37 \\
{\sysname} Adapter & 0.42M & 54.94 \\
{\sysname} LoRA & 0.35M & {\bf 55.64} \\
 \bottomrule
 
		\end{tabular}
	\end{center}
	\vspace*{-1em}
	\caption{Results on WebNLG with GPT-2 medium backbone. The results are based on all categories in the test set of WebNLG. Best result on each task is in \textbf{bold}. We report {\sysname} results with both adapters and LoRA as underlying PEFT method. {\sysname} outperforms all competing methods as well as fully fine-tuned large model with only $0.1\%$ tunable parameters.$^\dagger{}$~denotes results reported from \cite{hu2021lora} and repr. denotes reproduced results. \#Param. denotes the number of tunable adaptation parameters used during inference.}
	\label{tab:webnlg}
\vspace*{-1em}
\end{table}

\subsubsection{Few-shot NLU}

In contrast to the fully supervised setting in the above experiments, we also perform few-shot experiments on six GLUE tasks following the same setup (e.g., shots, train and test splits) and evaluation as in~\cite{wang2021list}. Detailed experimental configuration presented in Section~\ref{sec:few-shot} of Appendix. {\sysname} uses a mixture-of-adapters with prompt-based fine-tuning~\cite{gao2021making}.

\begin{table*}[h]

\centering

\small

 \begin{tabular}{lccccccc} 
	\toprule

 \textbf{Model} & {MNLI}  & {RTE}   & {QQP}&{SST2}& {Subj}& {MPQA} & \bf Avg. \\
 
\midrule

 Full Prompt Fine-tuning\textsuperscript{*}  &62.8 {\tiny(2.6)}  & 66.1 {\tiny(2.2)} &  71.1 {\tiny(1.5)}  & 91.5 {\tiny(1.0)}& 91.0 {\tiny(0.5)} &82.7 {\tiny(3.8)} & 77.5
 
 \\\midrule
 
 Head-only\textsuperscript{*} & 54.1 {\tiny(1.1)}  & 58.8 {\tiny(2.6)} & 56.7 {\tiny(4.5)} & 85.6 {\tiny(1.0)}& 82.1 {\tiny(2.5)} & 64.1 {\tiny(2.1)} &66.9
 
 \\
 
 BitFit\textsuperscript{*}&  54.4 {\tiny(1.3)} & 59.8 \tiny{(3.5)} & 58.6 {\tiny(4.4)}  & 87.3 {\tiny(1.1)} & 83.9 {\tiny(2.3)}& 65.8 {\tiny(1.8)}&68.3  \\
 
 
 Prompt-tuning\textsuperscript{*} & 47.3 {\tiny(0.2)}  & 53.0 {\tiny(0.6)} &39.9 {\tiny(0.7)}  & 75.7 {\tiny(1.7)} & 51.5 {\tiny(1.4)} & 70.9 {\tiny(2.4)} & 56.4
 
  \\

  Houlsby Adapter\textsuperscript{*}  & 35.7 {\tiny(1.1)} & 51.0 {\tiny(3.0)} & 62.8 {\tiny(3.0)}  & 57.0 {\tiny(6.2)} & 83.2 {\tiny(5.4)} & 57.2 {\tiny(3.5)} & 57.8

\\
 {LiST} Adapter\textsuperscript{*}  &{62.4 {\tiny(1.7)} } &{66.6 {\tiny(3.9)}}&{71.2 {\tiny(2.6)} }&{91.7 {\tiny(1.0)}} & {90.9 {\tiny(1.3)}} & {82.6 {\tiny(2.0)}} & 77.6
 \\
 	{{\sysname} Adapter} & \textbf{65.6 {\tiny(2.6)}}&\textbf{69.6 {\tiny(3.4)}}&\textbf{72.6 {\tiny(1.2)} }&\textbf{91.8 {\tiny(1.1)}} & \textbf{91.5 {\tiny(2.0)}} & \textbf{84.7 {\tiny(1.6)}} & \textbf{79.3}\\
 
 \bottomrule

\end{tabular}
\vspace*{-1em}
\caption{Average performance and standard deviation of several parameter-efficient fine-tuning strategies based on RoBERTa-large with $|\mathcal{K}| = 30$ training labels. The best performance is shown in \textbf{bold}. Prompt-tuning, Head-only and BitFit tune $1M$ model parameters during inference. Houlsby Adapter, LiST Adapter and AdaMix Adapter tune $14M$ model parameters. \textsuperscript{*} denotes that the results are taken from \cite{wang2021list}. }
\label{tab:few_shot}
\end{table*}

Table~\ref{tab:few_shot} shows the performance comparison among different PEFT methods with $|K| = 30$ labeled examples with RoBERTa-large as frozen encoder. We observe significant performance gap for most PEFT methods with full model prompt-based fine-tuning i.e. with all model parameters being updated. {\sysname} with adapters outperforms full model tuning performance for few-shot NLU similar to that in the fully supervised setting. Note that {\sysname} and LiST~\cite{wang2021list} use similar adapter design with prompt-based fine-tuning. 

\subsection{Ablation Study}
\label{sec:ablation}

We perform all the ablation analysis on {\sysname} with adapters for parameter-efficient fine-tuning.

\noindent\textbf{Analysis of adaptation merging.} In this ablation study, we do not merge adaptation modules and consider two different routing strategies at inference time: (a) randomly routing input to any adaptation module, and (b) fixed routing where we route all the input to the first adaptation module in {\sysname}. From Table~\ref{tab:ablation_merging}, we observe {\sysname} with adaptation merging to perform better than any of the other variants without the merging mechanism. Notably, all of the {\sysname} variants outperform full model tuning.


\begin{table}[!htb]

\small
	\begin{center}

		\begin{tabular}{p{5cm}p{1cm}p{0.5cm}}
			\toprule \bf Model    &\#Param.       &\bf Avg. \\ \midrule

			Full Fine-tuning&   110M &82.7\\ \midrule

				{\sysname} w/ Merging   &0.9M & \textbf{84.5}\\ 

				{\sysname} w/o Merging + RandomRouting  & 3.6M &83.3\\ 
{\sysname} w/o Merging + FixedRouting &0.9M &83.7 \\ 
{\sysname} w/o Merging + Ensemble &3.6M&83.2 \\ 


\bottomrule
			    
\end{tabular}
	\end{center}

	\caption{{\sysname} without adaptation merging and different routing and ensembling strategies. Average results are presented on GLUE development set with {BERT-base} encoder. Detailed task results in Table~\ref{tab:app_ablation_merging} of Appendix for BERT-base and RoBERTa-large encoders.
	}
	\label{tab:ablation_merging}

\end{table}

\begin{figure}

 	\centering
  	\includegraphics[width=0.8\linewidth]{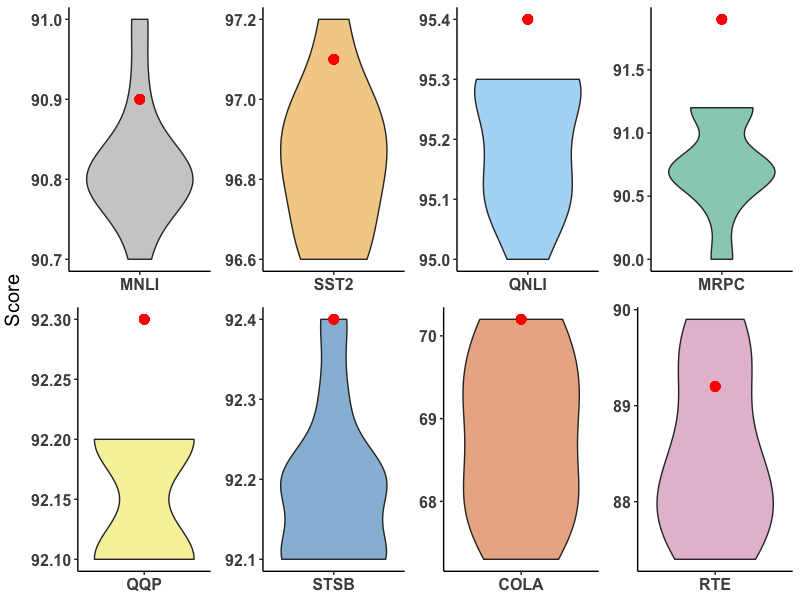}
 	  	\caption{Violin plot of {\sysname}-RandomRouting performance distribution with RoBERTa-large encoders. \textcolor{red}{Red dot} denotes the performance of {\sysname}. }
 	 	 	\label{fig:merging}
 
\end{figure}

Moreover, Figure~\ref{fig:merging} shows that the performance of merging mechanism is consistently better than the average performance of random routing and comparable to the best performance of random routing. 

\noindent\textbf{Averaging weights v.s. ensembling logits.} We compare {\sysname} with a variant of logit ensembling, denoted as {\sysname}-Ensemble. To this end, we make four random routing passes through the network for every input ($T$=$4$) and average the logits from different passes as the final predicted logit. Inference time for this ensembling method is 4 $\times$ {\sysname}. We run repeated experiments with three different seeds and report mean performance in Table~\ref{tab:ablation_merging}. We observe {\sysname} with adaptation weight averaging to outperform logit-ensembling following our analysis ($\mathcal{L}^{AM}_{\mathcal{W}}$ v.s. $\mathcal{L}^{Ens}_{\mathcal{W}}$) in Section~\ref{subsec:BNN}.



\noindent\textbf{Analysis of consistency regularization.} 
We drop consistency regularization during training for ablation and demonstrate significant performance degradation in Table~\ref{tab:ablation_design}.

\begin{table}[!htb]

\small
	\begin{center}
		\begin{tabular}{p{2.4cm}p{0.5cm}p{0.5cm}p{0.5cm}p{0.5cm}p{0.5cm}}
			\toprule \bf Model/\# Train&MNLI&QNLI&SST2&MRPC&RTE\\ 
			                            &393k                         &108k            &67k            &3.7k                     &2.5k             \\ 
			\midrule
			Full Fine-tuning & 90.2  & 94.7& 96.4&90.9&86.6 \\\midrule
				{\sysname}  & \textbf{90.9}&\textbf{95.4}&\textbf{97.1}&\textbf{91.9}& \textbf{89.2}\\

				
		    \hspace{0.4em}{w/o Consistency}    & 90.7&95.0&\textbf{97.1}&91.4&84.8\\ 
		    \hspace{0.4em}{w/o Sharing}   &\textbf{90.9} &95.0&96.4&90.4& 84.1\\

		    \bottomrule

		\end{tabular}
	\end{center}

	\caption{Ablation study demonstrating the impact of consistency regularization and sharing in {\sysname}.
	}	
	\label{tab:ablation_design}

\end{table}

\noindent\textbf{Analysis of adaptation module sharing.} 
We remove adaptation module sharing in {\sysname} for ablation and keep four different copies of project-down and four project-up FFN layers. From Table~\ref{tab:ablation_design} we observe the performance gap between {\sysname} and {\sysname} w/o sharing to increase with decrease in the dataset size demonstrating the importance of parameter sharing for low-resource tasks  (e.g., RTE, MRPC). This is further demonstrated in Figure~\ref{fig:training_loss} in Appendix which shows a faster convergence and lower training loss of {\sysname} with sharing  compared to that without given the same number of training steps. We explore which adaptation module to share (project-up v.s. project-down) in Table~\ref{tab:ablation_sharing_up_down} in Appendix that depict similar results.



\noindent\textbf{Impact of the number of adaptation modules.} In this study, we vary the number of adaptation modules in {\sysname} as $2$, $4$ and $8$ during training. Table~\ref{tab:ablation_adapter_num} shows diminishing returns on aggregate task performance with increasing number of modules. 
As we increase sparsity and the number of tunable parameters by increasing the number of adaptation modules, low-resource tasks like RTE and SST-2 -- with limited amount of labeled data for fine-tuning -- degrade in performance compared to high-resource tasks like MNLI and QNLI.

\begin{table}[!htb]

\small
	\begin{center}
		\begin{tabular}{p{1.4cm}p{0.6cm}p{0.6cm}p{0.6cm}p{0.6cm}p{0.6cm}}
			\toprule  Adaptation        &MNLI                             &QNLI    &SST2       &MRPC                   &RTE           \\ 
			           Module               &393k                         &108k            &67k            &3.7k                     &2.5k              \\ \midrule

			2   & \bf 90.9 &95.2& 96.8&90.9&87.4\\ 
		    {4}\textsuperscript{*}   &  \bf 90.9 & \bf 95.4& \bf 97.1& \bf 91.9&  \bf89.2\\
		    8   & \bf 90.9& 95.3& 96.9&91.4&87.4\\
		    		    \bottomrule

		\end{tabular}
	\end{center}

	\caption{Varying the number of adaptation modules in {\sysname} with RoBERTa-large encoder. \textsuperscript{*} denotes the number of modules  used in {\sysname} with adapters.
	}
	\label{tab:ablation_adapter_num}

\end{table}

\noindent\textbf{Impact of adapter bottleneck dimension.} 
Table~\ref{tab:ablation_adapter_dim} shows the impact of bottleneck dimension of adapters with different encoders in {\sysname}. The model performance improves with increase in the number of trainable parameters by increasing the bottleneck dimension with diminishing returns after a certain point.

\begin{table}[!htb]

\small
	\begin{center}
		\begin{tabular}{p{1cm}p{0.8cm}p{0.6cm}p{0.6cm}p{0.6cm}p{0.6cm}p{0.6cm}}
		\toprule \bf Adapter         &\#Param.    &MNLI                             &QNLI     &SST2       &MRPC                   &RTE          \\ 
			                             Dimension &&393k                         &108k            &67k            &3.7k                     &2.5k               \\ 
			                             \midrule 
			8   & 0.4M& 90.7  &95.2& 96.8&91.2&87.7\\ 
		    16\textsuperscript{*}  & 0.8M & 90.9& \bf 95.4& \bf 97.1& \bf 91.9& \bf 89.2\\ 
		    32   & 1.5M & \bf 91.0& \bf 95.4&96.8&90.7& \bf 89.2\\ \bottomrule

		\end{tabular}
	\end{center}

	\caption{Varying the bottleneck dimension of adapters in {\sysname} with RoBERTa-large encoder. \textsuperscript{*} denotes the bottleneck dimension used in {\sysname} with adapters. Results with BERT-base encoder in Table~\ref{tab:app-ablation_adapter_dim} in Appendix.
	}
	\label{tab:ablation_adapter_dim}


\end{table}

\section{Related Work}

\noindent\textbf{Parameter-efficient fine-tuning of PLMs.} Recent works on parameter-efficient fine-tuning (PEFT) can be roughly categorized into two categories: (1) tuning a subset of existing parameters including head fine-tuning~\cite{lee2019would}, bias term tuning~\cite{zaken2021bitfit}, (2) tuning newly-introduced parameters including adapters~\cite{houlsby, pfeiffer2020AdapterHub}, prompt-tuning~\cite{prompt_tuning},  prefix-tuning~\cite{prefix} and low-rank adaptation~\cite{hu2021lora}. As opposed to prior works operating on a single adaptation module, {\sysname} introduces a mixture of adaptation modules with stochastic routing during training and adaptation module merging during inference to keep the same computational cost as with a single module. Further, {\sysname} can be used on top of any PEFT method to further boost its performance.

\noindent\textbf{Mixture-of-Expert (MoE).} ~\citealp{shazeer2017outrageously} introduced the MoE model with a single gating network with $Top$-$k$ routing and load balancing across experts.~\citealp{fedus2021switch} propose initialization and training schemes for $Top$-$1$ routing.~\citealp{zuo2021taming} propose consistency regularization for random routing;~\citealp{yang2021exploring} propose $k$ $Top$-$1$ routing with expert-prototypes, and~\citealp{Roller2021HashLF,Lewis2021BASELS} address other load balancing issues. All the above works study sparse MoE with pre-training the entire model from scratch. In contrast, we study parameter-efficient adaptation of pre-trained language models by tuning only a very small number of sparse adapter parameters.

\noindent\textbf{Averaging model weights.} Recent explorations~\cite{szegedy2016rethinking,matena2021merging, modelsoup, izmailov2018averaging} study model aggregation by averaging all the model weights. \cite{matena2021merging} propose to merge pre-trained language models which are fine-tuned on various text classification tasks. \cite{modelsoup} explores averaging model weights from various independent runs on the same task with different hyper-parameter configurations. In contrast to the above works on full model fine-tuning, we focus on parameter-efficient fine-tuning. We explore weight averaging for merging weights of adaptation modules consisting of small tunable parameters that are updated during model tuning while keeping the large model parameters fixed.

\section{Conclusions}

We develop a new framework {\sysname} for parameter-efficient fine-tuning (PEFT) of large pre-trained language models (PLM). {\sysname} leverages a mixture of adaptation modules to improve downstream task performance without increasing the computational cost (e.g., FLOPs, parameters) of the underlying adaptation method. We demonstrate {\sysname} to work with and improve over different PEFT methods like adapters and low rank decompositions across NLU and NLG tasks. 

By tuning only $0.1-0.2\%$ of PLM parameters, {\sysname} outperforms full model fine-tuning that updates all the model parameters as well as other state-of-the-art PEFT methods.

\section{Limitations}
The proposed {\sysname} method is somewhat compute-intensive as it involves fine-tuning large-scale language models. The training cost of the proposed AdaMix is higher than standard PEFT methods since the training procedure involves multiple copies of adapters. Based on our empirical observation, the number of training iterations for AdaMix is usually between 1$\sim$2 times the training for standard PEFT methods.  This imposes negative impact on carbon footprint from training the described models. 

{\sysname} is orthogonal to most of the existing parameter-efficient fine-tuning  (PEFT) studies and is able to potentially improve the performance of any PEFT method. In this work, we explore two representative PEFT methods like adapter and LoRA but we did not experiment with other combinations like prompt-tuning and prefix-tuning. We leave those studies to future work.

\section{Acknowledgment}

The authors would like to thank the anonymous referees for their valuable comments and helpful suggestions and would like to thank Guoqing Zheng and Ruya Kang for their insightful comments on the project. This work is supported in part by the US National Science Foundation under grants NSF-IIS 1747614 and NSF-IIS-2141037. Any opinions, findings, and conclusions or recommendations expressed in this material are those of the author(s) and do not necessarily reflect the views of the National Science Foundation.

\bibliography{anthology,custom,iclr2022_conference}
\bibliographystyle{acl_natbib}

\clearpage
\section*{Appendix}
\label{sec:appendix}

\appendix

 \section{Few-shot NLU Datasets}
 \label{sec:few-shot}
 \textbf{Data.} In contrast to the fully supervised setting in the above experiments, we also perform few-shot experiments following the prior study~\cite{wang2021list} on six tasks including  MNLI~\cite{williams2018broad-mnli}, RTE~\citep{dagan2005pascal-rte1,bar2006second,giampiccolo2007third-rte3,bentivogli2009fifth-rte4}, QQP\footnote{\url{https://www.quora.com/q/quoradata/}} and SST-2~\citep{socher2013recursive-sst-2}. The results are reported on their development set following~\citep{zhang2020revisiting}.  MPQA~\citep{wiebe2005annotating} and Subj~\citep{pang2004sentimental-subj} are used for polarity and subjectivity detection, where we follow~\cite{gao2021making} to keep $2,000$ examples for testing. The few-shot model only has access to $|\mathcal{K}|$ labeled samples for any task.  Following \textit{true few-shot learning} setting~\citep{perez2021true, wang2021list}, we {\em do not use any additional validation set} for any hyper-parameter tuning or early stopping. The performance of each model is reported after fixed number of training epochs. For a fair comparison, we use the same set of few-shot labeled instances for training as in \cite{wang2021list}. We train each model with $5$ different seeds and report average performance with standard deviation across the runs. In the few-shot experiments, we follow \cite{wang2021list} to train {\sysname} via the prompt-based fine-tuning strategy. In contrast to \cite{wang2021list}, we do not use any unlabeled data. 
 
\section{Ablation Study}
\label{appendix_sec:ablation}

\begin{table}[htb]
\small
	\begin{center}
		\begin{tabular}{lcc}
			\toprule \bf Model            &MNLI                 &SST2          \\ 
			                            &Acc                                               &Acc               \\ \midrule
			\midrule
			Sharing Project-up   & 90.9&97.1\\ \midrule
		   
		    Sharing Project-down  &90.8 & 97.1\\ \bottomrule

		\end{tabular}
	\end{center}
	\caption{Ablation study demonstrating the impact of parameter sharing in {\sysname} adapter framework.
	}
	\label{tab:ablation_sharing_up_down}
\end{table}

\begin{table}[htb]
\small
	\begin{center}
	\resizebox{\columnwidth}{!}{%
		\begin{tabular}{ccccccc}
		\toprule \bf Adapter Dim        & \#Param.    &MNLI                             &QNLI     &SST2       &MRPC                   &RTE          \\ 
			                           &  &393k                         &108k            &67k            &3.7k                     &2.5k               \\ 
			                             \midrule \midrule
			 \multicolumn{7}{c}{\textbf{BERT\textsubscript{BASE}}}\\ \midrule	
			            8 & 0.1M &82.2 & 91.1& 92.2&87.3&72.6\\ 	
		    16  &0.3M & 83.0 &91.5&92.2&88.2&72.9 \\ 	
		    32 & 0.6M  & 83.6  & 91.3&92.2&88.5&73.6 \\
		    	48\textsuperscript{*} & 0.9M & \bf 84.7&91.5& \bf 92.4& \bf 89.5& 74.7\\ 
		    64 &1.2M &84.4& \bf 91.8& 92.3&88.2& \bf 75.1\\\midrule
		    	 \multicolumn{7}{c}{\textbf{RoBERTa\textsubscript{LARGE}}}\\ \midrule	
			8   & 0.4M & 90.7  &95.2& 96.8&91.2&87.7\\ 
		    16\textsuperscript{*} &0.8M  & 90.9& \bf 95.4& \bf 97.1& \bf 91.9& \bf 89.2\\ 
		    32   & 1.5M &\bf 91.0& \bf 95.4&96.8&90.7& \bf 89.2\\ \bottomrule

		\end{tabular}
		}
	\end{center}
	\caption{Varying the bottleneck dimension of adapters in {\sysname} with BERT-base and RoBERTa-large encoder. \textsuperscript{*} denotes the bottleneck dimension used in {\sysname} with adapters.
	}
	\label{tab:app-ablation_adapter_dim}
\end{table}

\section{Detailed Results on NLU Tasks}

The results on NLU tasks are included in Table~\ref{tab:roberta_glue_dev} and Table~\ref{tab:app-bert_glue_dev}. The performance {{\sysname} with RoBERTa-large encoder achieves the best performance in terms of different task metrics in the GLUE benchmark.
 {\sysname} with adapters is the only PEFT method which outperforms full model fine-tuning on all the tasks and on average score.}
 Additionally, the improvement brought by  {\sysname} is more significant with BERT-base as the encoder, demonstrating 2.2\% and 1.2\% improvement over the performance of full model fine-tuning and the best performing baseline UNIPELT with BERT-base. The improvement is observed to be consistent as that with RoBERTa-large on every task. 
 The NLG results are included in Table~\ref{tab:dart} and \ref{tab:webnlg}.

\begin{table*}[htb]
\small
	\begin{center}
		\begin{tabular}{lrccccccc  c c}
			\toprule \bf Model    &\#Param.       &MNLI        &QNLI          &SST2              &QQP          &MRPC               &CoLA             &RTE           &STS-B&\bf Avg. \\ 
			                            &&Acc             &Acc             &Acc          &Acc /F1           &Acc/F1             &Mcc            &Acc            &Pearson     \\ 
			\midrule
			Full Fine-tuning$^\dagger{}$&   110M &83.2&90.0 &91.6& -/87.4&-/90.9&62.1&66.4&89.8&82.7\\ 
			Houlsby Adapter$^\dagger{}$&0.9M& 83.1&90.6&91.9&-/86.8&-/89.9&61.5&71.8&88.6&83.0\\ 
			 BitFit$^\diamond$ & 0.1M& 81.4&90.2&92.1&-/84.0&-/90.4&58.8&72.3&89.2& 82.3\\
			Prefix-tuning$^\dagger{}$&0.2M &81.2&90.4&90.9&-/83.3&-/91.3&55.4&\textbf{76.9}&87.2&82.1\\ 
			LoRA$^\dagger{}$&0.3M  &82.5&89.9&91.5&-/86.0&-/90.0&60.5&71.5&85.7&82.2\\ 
			UNIPELT (AP)$^\dagger{}$&1.1M&83.4&90.8&91.9&-/86.7&-/90.3&61.2&71.8&88.9&83.1\\ 
			UNIPELT (APL)$^\dagger{}$&1.4M&83.9&90.5&91.5&85.5&-/90.2&58.6&73.7&88.9&83.5\\ \midrule

				{{\sysname} Adapter}  &0.9M &\textbf{84.7}&\textbf{91.5}&\textbf{92.4}&\textbf{90.7/}&\textbf{89.5/}&\textbf{62.9}&74.7&\textbf{89.9}& \textbf{84.5}\\
								  &0.9M &\textbf{}&\textbf{}&\textbf{}&\textbf{87.6}&\textbf{92.4}&\textbf{}&&\textbf{}& \textbf{}\\
				\bottomrule

		\end{tabular}
	\end{center}
	\caption{Main results on GLUE development set with \textbf{BERT-base} encoder. The best result on each task is in \textbf{bold} and ``-'' denotes the missing measure. $^\dagger{}$ and $^\diamond$ denote that the reported results are taken from \cite{mao2021unipelt,zaken2021bitfit}. The average performance is calculated based on F1 of QQP and MRPC. \#Param. refers to the number of updated parameters in the inference stage.
	}
	\label{tab:app-bert_glue_dev}
\end{table*}

\begin{figure*}[hbt]
\centering
\subfloat[BERT-base]{
\begin{minipage}{0.5\linewidth}
\includegraphics[width=0.9\linewidth]{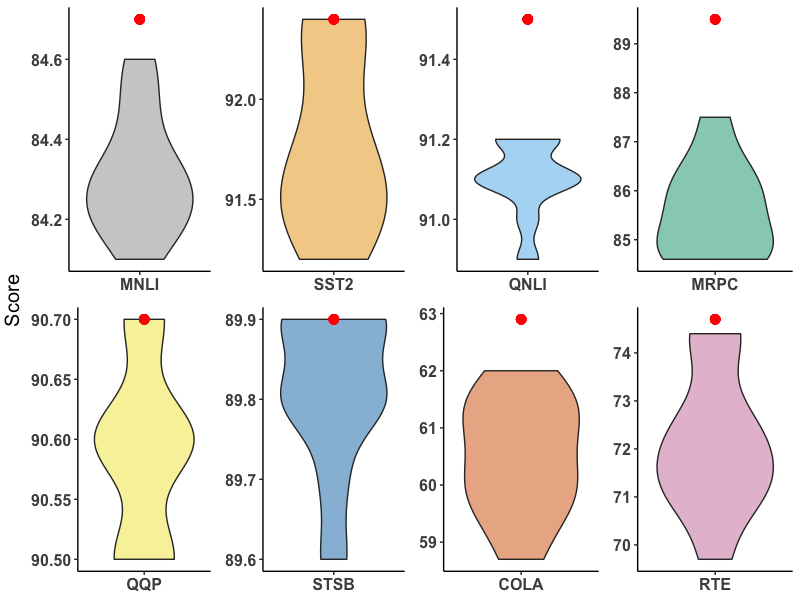}
\label{subfig:bert}
\end{minipage}}
\subfloat[RoBERTa-large]{
\begin{minipage}{0.5\linewidth}
\includegraphics[width=0.9\linewidth]{figs/roberta_large.png}
\label{subfig:}
\end{minipage}}
 \caption{Violin plot of {\sysname}-RandomRouting performance distribution with BERT-base and RoBERTa-large encoders. \textcolor{red}{Red dot} denotes the performance of {\sysname}. }\label{fig:random_routing}
\end{figure*}

\begin{table*}[!htb]
\small
	\begin{center}
		\begin{tabular}{lrccccccccc}
			\toprule \bf Model    &\#Param.       &MNLI        &QNLI          &SST2              &QQP          &MRPC               &CoLA             &RTE           &STS-B&\bf Avg. \\ 
			                            &&Acc             &Acc             &Acc          &Acc /F1           &Acc/F1             &Mcc            &Acc            &Pearson  \\\midrule \midrule
            \multicolumn{11}{c}{\textbf{BERT\textsubscript{BASE}}}\\ \midrule	
			Full Fine-tuning&   110M &83.2&90.0 &91.6& -/87.4&-/90.9&62.1&66.4&89.8&82.7\\ \midrule

			{\cellcolor{na}	{\sysname}}  &0.9M &\textbf{84.7}&\textbf{91.5}&\textbf{92.4}&\textbf{90.7/}&\textbf{89.5/}&\textbf{62.9}&\textbf{74.7}&\textbf{89.9}& \textbf{84.5}\\ 
						{}  & &\textbf{}&\textbf{}&\textbf{}&\textbf{87.6}&\textbf{92.4}&\textbf{}&\textbf{}&\textbf{}& \textbf{}\\
			\midrule
				{\sysname}-RandomRouting  & 3.6M &84.3&91.1&91.8&90.6/&85.6/&60.5&72.1&89.8&83.3\\ 
								&  &&&&87.4&89.1&&&&\\
				\midrule
{\sysname}-FixedRouting &0.9M & 84.5 & 91.1 & 91.6&90.5/&87.5/&61.4&73.3&89.8&83.7 \\ 
 & &  &  & &87.3&90.8&&&& \\
\midrule
{\sysname}-Ensemble &3.6M& 84.3&91.2&91.6&90.5/&85.9/&59.4&72.1&89.8&83.2 \\ 
 && &&&87.4&89.4&&&& \\
\midrule

\multicolumn{11}{c}{\textbf{RoBERTa\textsubscript{LARGE}}}\\ \midrule	
Full Fine-tuning &355.0M&90.2&94.7&96.4& 92.2/-&90.9/-&68.0&86.6&\textbf{92.4}&88.9 \\ \midrule

	{\cellcolor{na}{\sysname}}&0.8M &\textbf{90.9}&\textbf{95.4}&\textbf{97.1}&\textbf{92.3/}&\textbf{91.9/}&\textbf{70.2}&\textbf{89.2}&\textbf{92.4} &\textbf{89.9}\\ 
& &\textbf{}&\textbf{}&\textbf{}&\textbf{89.8}&\textbf{94.1}&\textbf{}&\textbf{}&\textbf{} &\textbf{}\\
\midrule
{\sysname}-RandomRouting  &3.2M & 90.8&95.2&96.8&92.2/&90.8/&68.8&88.5&92.2&89.4\\ 
  &&&&&89.6&93.3&&&&\\
\midrule
{\sysname}-FixedRouting &0.8M &90.7&95.1&96.8&92.1/&91.2/&68.6&\bf 89.2&92.2&89.5\\
& &&&&89.5&93.6&&&&\\
\midrule
{\sysname}-Ensemble &3.2M &\bf 90.9&95.3&97.0&92.2/&91.0/&69.3&89.1&\bf 92.4&89.7\\
 &&&&&89.7&93.5&&&&\\
\bottomrule
			    
\end{tabular}
	\end{center}
	\caption{Comparing the impact of different routing and ensembling strategies with {\sysname}. Results are presented on GLUE development set with {BERT-base} and RoBERTa-large encoders. Average results are calculated following Table~\ref{tab:roberta_glue_dev} and Table~\ref{tab:bert_glue_dev} for consistency. The best result on each task is in \textbf{bold} and ``-'' denotes the missing measure.
	}
	\label{tab:app_ablation_merging}

\end{table*}

\begin{figure*}[!hbt]
\centering
\subfloat[MNLI]{
\begin{minipage}{0.33\linewidth}
\includegraphics[width=0.9\linewidth]{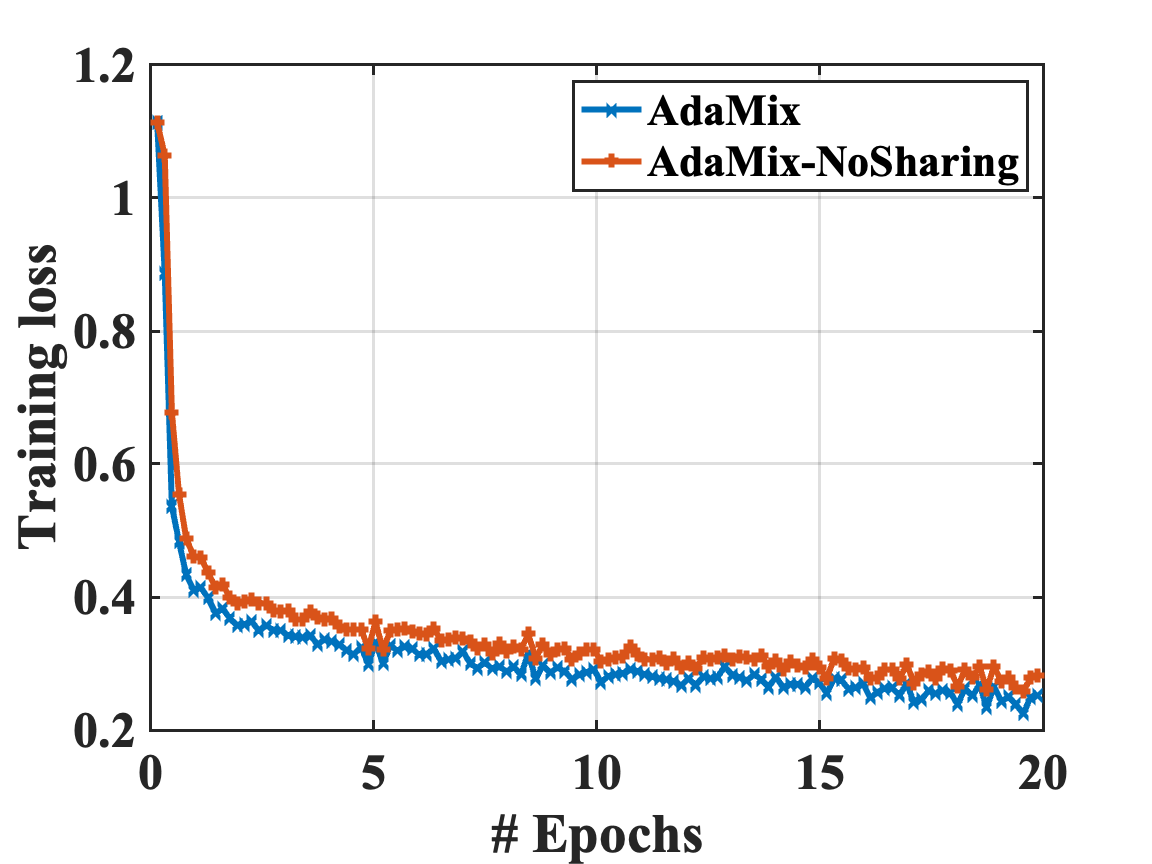}
\label{subfig:mnli}
\end{minipage}}
\subfloat[QNLI]{
\begin{minipage}{0.33\linewidth}
\includegraphics[width=0.9\linewidth]{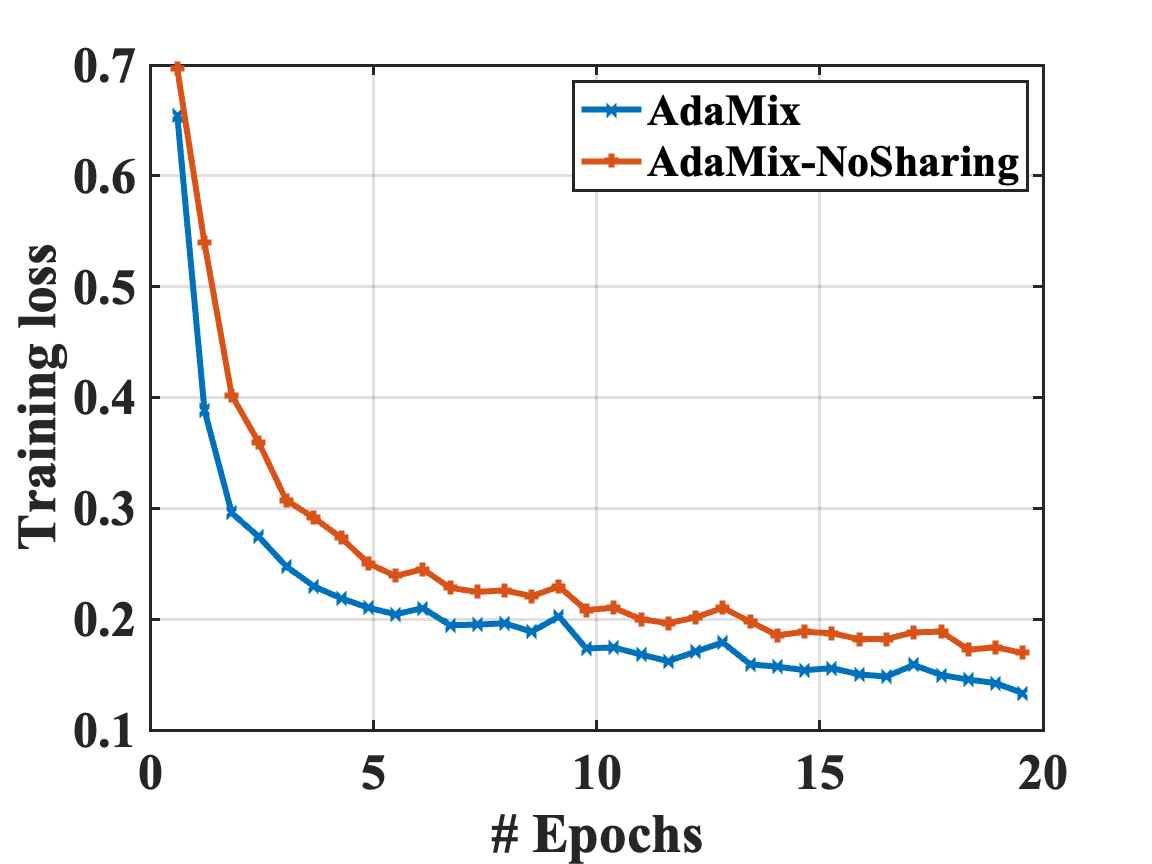}
\label{subfig:qnli}
\end{minipage}}
\subfloat[SST2]{
\begin{minipage}{0.33\linewidth}
\includegraphics[width=0.9\linewidth]{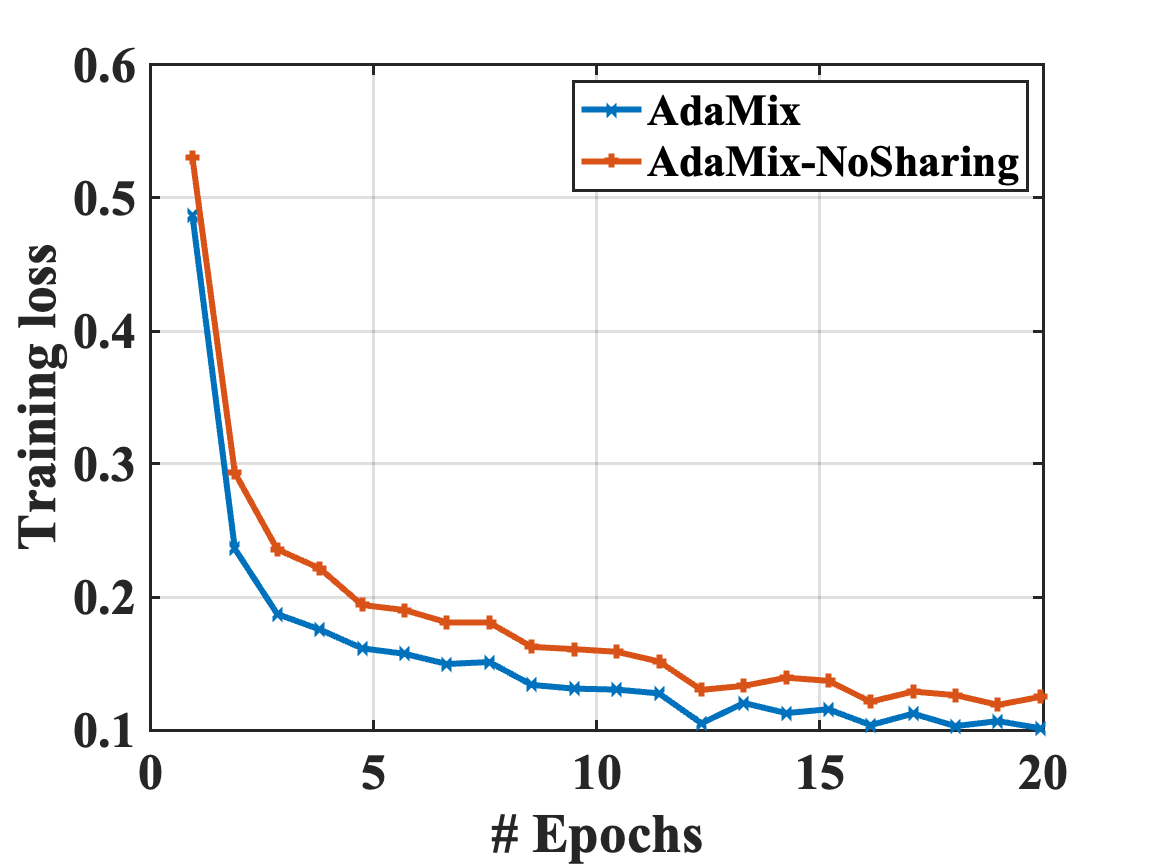}
\label{subfig:sst}
\end{minipage}}
 \caption{Convergence analysis demonstrating the impact of adapter sharing design in {\sysname}. }\label{fig:training_loss}
\end{figure*}

\section{Hyper-parameter}
\label{sec:hyper}
Detailed hyper-parameter configuration for different tasks presented in Table~\ref{tab:hyper-glue_dev} and Table~\ref{tab:hyper-nlg}.

\begin{table*}
\small
	\begin{center}
		\begin{tabular}{@{\hskip1pt}l@{\hskip1pt}|@{\hskip1pt}c@{\hskip1pt}|c@{\hskip1pt}|c@{\hskip1pt}|c@{\hskip1pt}|c@{\hskip1pt}|@{\hskip1pt}c @{\hskip1pt}|@{\hskip1pt} c @{\hskip1pt}}
			\toprule \bf Task & Learning rate & epoch       &batch size        &warmup                &weight decay  & adapter size & adapter num    \\ \midrule \bottomrule
	
	\multicolumn{8}{c}{\textbf{BERT\textsubscript{BASE}}}\\ \midrule	
	
           MRPC& 4e$-4$ & 100 & 16 & 0.06 & 0.1 & 48 & 4\\
           CoLA&5e$-4$&100&16& 0.06 & 0.1 & 48 & 4\\
           SST&4e-4&40&64& 0.06 & 0.1 & 48 & 4\\
           STS-B&5e-4&80&32& 0.06 & 0.1 & 48 & 4\\
           QNLI&4e-4&20&64& 0.06 & 0.1 & 48 & 4\\
           MNLI&4e-4&40&64& 0.06 & 0.1 & 48 & 4\\
           QQP&5e-4&60&64& 0.06 & 0.1 & 48 & 4\\
           RTE&5e-4&80&64& 0.06 & 0.1 & 48 & 4\\
           	\midrule
           	\multicolumn{8}{c}{\textbf{RoBERTa\textsubscript{LARGE}}}\\ \midrule	
                 MRPC&3e-4&60&64&0.6&0.1&16&4\\
           CoLA&3e-4&80&64&0.6&0.1&16&4\\
           SST&3e-4&20&64&0.6&0.1&16&4\\
           STS-B&3e-4&80&64&0.6&0.1&16&4\\
           QNLI&3e-4&20&64&0.6&0.1&16&4\\
           MNLI&3e-4&20&64&0.6&0.1&16&4\\
           QQP&5e-4&80&64&0.6&0.1&16&4\\
           RTE&5e-4&60&64&0.6&0.1&16&4\\
               	\bottomrule
        
		\end{tabular}
	\end{center}
	\caption{Hyperparameter configurations for GLUE tasks.}
	\label{tab:hyper-glue_dev}
\end{table*}

\begin{table*}
\small
	\begin{center}
		\begin{tabular}{@{\hskip1pt}l@{\hskip1pt}|@{\hskip1pt}c@{\hskip1pt}|c@{\hskip1pt}|@{\hskip1pt}c @{\hskip1pt}|@{\hskip1pt} c @{\hskip1pt}}
			\toprule \bf Task & epoch        & warmup steps & adapter size & no. of experts    \\ \midrule \bottomrule

    \multicolumn{5}{c}{\textbf{Adapter with Adamix}}\\ \midrule
	
           E2E NLG Challenge & 20 & 2000 & 8 & 8\\
           WebNLG & 25 & 2500 & 8 & 8\\
           DART & 20 & 2000 & 8 & 8\\
           \midrule
           	\multicolumn{5}{c}{\textbf{LoRA with Adamix}}\\ \midrule	
           E2E NLG Challenge & 20 & 2000 & $-$ & 8\\
           WebNLG & 25 & 2500 & $-$ & 8\\
           DART & 20 & 2000 & $-$ & 8\\
           \bottomrule
        
		\end{tabular}
	\end{center}
	\caption{Hyperparameter configurations for GPT-2 \textsubscript{Medium} on NLG tasks. We retain all other default training and generation specific hyper-parameters from LoRA~\cite{hu2021lora}.}
	\label{tab:hyper-nlg}
\end{table*}


\end{document}